# Federated Anomaly Detection for Multi-Tenant Cloud Platforms with Personalized Modeling

Yuxi Wang
Carnegie Mellon University
Pittsburgh, USA

Heyao Liu
Northeastern University
Boston, USA

Nyutian Long
New York University
New York, USA

Guanzi Yao*
Northwestern University
Evanston, USA

*Abstract*-This paper proposes an anomaly detection method based on federated learning to address key challenges in multi-tenant cloud environments, including data privacy leakage, heterogeneous resource behavior, and the limitations of centralized modeling. The method establishes a federated training framework involving multiple tenants. Each tenant trains the model locally using private resource usage data. Through parameter aggregation, a global model is optimized, enabling cross-tenant collaborative anomaly detection while preserving data privacy. To improve adaptability to diverse resource usage patterns, a personalized parameter adjustment mechanism is introduced. This allows the model to retain tenant-specific feature representations while sharing global knowledge. In the model output stage, the Mahalanobis distance is used to compute anomaly scores. This enhances both the accuracy and stability of anomaly detection. The experiments use real telemetry data from a cloud platform to construct a simulated multi-tenant environment. The study evaluates the model's performance under varying participation rates and noise injection levels. These comparisons demonstrate the proposed method's robustness and detection accuracy. Experimental results show that the proposed method outperforms existing mainstream models across key metrics such as Precision, Recall, and F1-Score. It also maintains stable performance in various complex scenarios. These findings highlight the method's practical potential for intelligent resource monitoring and anomaly diagnosis in cloud computing environments.

*Keywords-Multi-tenant cloud computing; federated anomaly detection; model personalization; robustness analysis*

I. INTRODUCTION

With the rapid growth of cloud computing, muti-tenant which plays an essential role for resource usage optimization, also introduces complex issues such as resource contention, performance interference, and security risks [1]. These challenges are particularly evident in anomaly detection and resource state management. Traditional centralized detection mechanisms often struggle with performance bottlenecks, privacy risks, and inadequate sensitivity. Therefore, we propose a federated learning-based anomaly detection method that addresses the challenges [2].

In multi-tenant cloud environments, resource usage is highly dynamic, complex in patterns, and frequently changing[3]. Tenants often influence each other's resource usage behaviors. In scenarios with high concurrency and elastic scaling, anomalies such as sudden CPU spikes, memory leaks, and network congestion occur frequently[4]. Traditional detection methods mostly rely on predefined rules or static thresholds. These are inadequate for capturing hidden and evolving abnormal behaviors. Moreover, cloud platforms handle large volumes of sensitive data and user privacy. This leads to the problem of data silos, making it difficult to train high-quality models without exposing private data[5]. Hence, it is essential to develop an intelligent detection method that protects data privacy while enabling distributed learning.

Federated learning, as a recently emerging distributed machine learning framework, offers a novel approach for intelligent detection in cloud computing environments. Its core advantage lies in training models locally without transferring raw data. This effectively reduces the risk of privacy leakage and relieves the pressure on data transmission. It is particularly suitable for sensitive scenarios where tenants in the cloud cannot share raw data. By training models locally and aggregating parameter updates, federated learning supports a detection mechanism that balances global modeling and local customization[6]. This enables more accurate identification of various resource anomalies. Compared with centralized methods, it offers better scalability and real-time performance. It is also more effective in handling data heterogeneity and diverse resource usage patterns.

In addition, federated learning has shown increasing advantages in analyzing cross-tenant resource behaviors. In multi-tenant environments, business types, resource demands, and access patterns vary significantly among tenants. A single model often fails to generalize across all scenarios. Federated learning enables knowledge transfer through cross-tenant modeling. This helps build more robust anomaly detection systems. Without centralized data storage, the platform can still obtain comprehensive behavioral feature representations. This supports the timely detection of system-wide fault trends, malicious activity patterns, and potential resource bottlenecks. It provides technical support for cloud service providers to improve overall service quality and platform stability[7].

Building a resource anomaly detection system for multi-tenant cloud platforms based on federated learning holds significant technical value and broad application prospects [8-11]. Technically, it addresses key challenges in distributed intelligent monitoring, such as data security, model generalization, and system complexity. From a practical perspective, it helps improve cloud platform operations, reduce anomaly handling costs, and enhance service stability and user experience. As cloud computing continues to scale and resource scheduling becomes more intelligent, developing anomaly detection mechanisms integrated with federated learning has become a crucial direction for advancing intelligent cloud evolution.

## II. RELATED WORK

The challenge of robust anomaly detection in multi-tenant cloud environments intersects with recent advances in federated learning, sequential modeling, model adaptation, and resource optimization. For example, Y. Xing et al. present time-aware and multi-channel convolutional user modeling for sequential recommendation, showing that capturing dynamic and heterogeneous user behaviors through advanced sequence modeling techniques is crucial for effective personalization and temporal anomaly detection in large-scale systems [12]. Their methodology provides inspiration for our model's handling of diverse resource usage patterns across tenants.

X. Sun et al. introduce a dynamic scheduling framework using double DQN reinforcement learning, demonstrating how distributed, policy-driven optimization can effectively handle resource contention and system adaptation under dynamic workloads [13]. This approach underscores the benefits of decentralized training and online adaptation, which are foundational to federated anomaly detection in privacy-sensitive cloud settings. J. Liu et al. apply improved A3C reinforcement learning to market turbulence prediction and risk control, emphasizing the value of real-time, adaptive policy adjustment in environments with high variability and uncertainty [14]. This concept translates directly to our framework's ability to maintain robustness and performance stability amidst fluctuating resource demands and noisy data.

For distributed resource management, Y. Zou et al. propose autonomous resource orchestration in microservices using reinforcement learning. Their method leverages localized feedback and cross-entity collaboration, which aligns closely with our federated parameter aggregation and personalized adaptation strategies for handling tenant-specific behaviors while enabling global knowledge transfer [15].

Complex data dependencies and high-dimensional feature spaces are further addressed by N. Jiang et al., who combine graph convolution and sequential modeling for scalable network traffic estimation. Their joint spatial-temporal modeling directly inspires our unified anomaly representation and the integration of diverse behavioral signals across distributed tenants [16]. Deep probabilistic models, such as the mixture density networks for anomaly detection described by L. Dai et al., enable robust, distribution-aware identification of rare and evolving anomalies in user behaviors [17]. This probabilistic perspective is reflected in our use of Mahalanobis distance for scoring and robust outlier detection in multi-tenant scenarios.

Context and semantics also play critical roles. J. Liu et al. develop a dynamic transformer-based framework for context-aware rule mining, showing how deep contextual encoding can improve sensitivity to evolving rules and hidden anomalies in complex environments [18]. Similarly, Y. Cheng et al. propose a hybrid deep learning architecture integrating CNN and BiLSTM for financial risk prediction, demonstrating that multi-layer and multi-perspective feature extraction enhances model generalization—a principle echoed in our federated learning pipeline [19]. Finally, the structuring of low-rank adaptation with semantic guidance, as explored by H. Zheng et al., provides new perspectives on personalized model fine-tuning. Their approach supports efficient adaptation to tenant-specific resource behaviors while preserving global consistency, directly informing our personalized parameter adjustment mechanism within the federated framework [20].

Collectively, these studies provide the methodological foundations for our federated, personalized, and robust anomaly detection approach, enabling adaptive and privacy-preserving monitoring across complex multi-tenant cloud platforms.

## III. METHOD

This study proposes a cloud computing multi-tenant resource anomaly detection method based on federated learning, aiming to achieve efficient cross-tenant anomaly identification while protecting data privacy [21]. The method uses federated learning as the basic framework, trains local models on each tenant node separately, and collaboratively optimizes the global detection model without sharing the original data. The model architecture is shown in Figure 1.

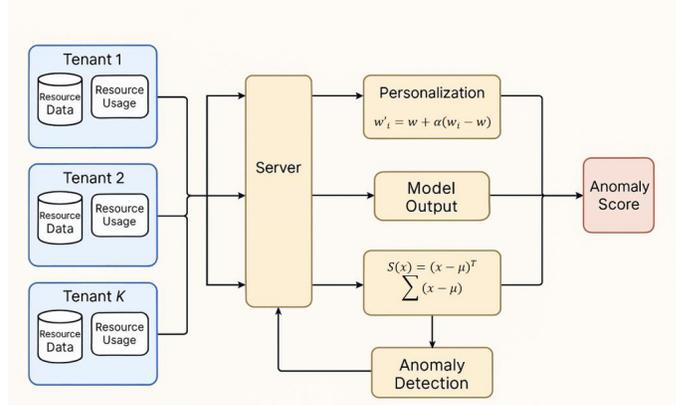

Figure 1. Overall model architecture diagram

In each round of federated training, tenant local node i uses its private resources to use the dataset $D_i = \{(x_j^{(i)}, y_j^{(i)})\}_{j=1}^{n_i}$ to optimize the local model parameters $w_i$. The objective function is defined as:

$$\min_{w_i} \frac{1}{n_i} \sum_{j=1}^{n_i} L(f(x_j^{(i)}; w_i), y_j^{(i)}) \qquad (1)$$

Among them, $L(\cdot)$ represents the loss function, which can usually be cross-entropy loss or mean square error, and $f(\cdot; w_i)$ represents the predicted output of the local model. To prevent the model from overfitting or gradient explosion, the L2 regularization term is introduced, and the optimization objective is further expressed as:

$$\min_{w_i} \frac{1}{n_i} \sum_{j=1}^{n_i} L(f(x_j^{(i)}; w_i), y_j^{(i)}) + \lambda \| w_i \|^2 \quad (2)$$

After each node completes local training, it uploads the model parameters to the central server for aggregation. The aggregation process uses the classic federated average algorithm (FedAvg), which synthesizes the global model parameters $w$ by weighted average:

$$w = \sum_{i=1}^{K} \frac{n_i}{n} w_i \quad (3)$$

K is the total number of tenant nodes participating in the training and $n = \sum_{i=1}^{K} n_i$ is the total number of samples of all nodes. The aggregated global model will be redistributed to each tenant node for the next round of local training iterations, thereby continuously improving the model's anomaly detection capabilities.

In order to improve the sensitivity and generalization ability of anomaly detection, an anomaly measurement mechanism based on Mahalanobis distance is introduced in the model output stage. Specifically, let the feature mean vector of each tenant resource state in the training set be $\mu$, and the covariance matrix be $\Sigma$. Then the anomaly score $S(x)$ of the feature input $x$ at a certain moment is defined as:

$$s(x) = (x - \mu)^T \Sigma^{-1} (x - \mu) \quad (4)$$

This anomaly score can be used as a reference indicator for the model to determine whether it is abnormal or not. In order to adapt to the cloud environment with dynamically changing resources, the above distribution parameters $\mu$ and $\Sigma$ can be continuously updated through the sliding window mechanism to maintain the model's sensitivity to new behavior patterns. At the same time, this study considers the heterogeneity of tenant data and introduces a personalized sub-model adjustment mechanism in some scenarios, so that the global model parameters can be lightly fine-tuned according to tenant characteristics after aggregation. The formula is as follows:

$$w'_i = w + \alpha (w_i - w) \quad (5)$$

$\alpha \in [0,1]$ controls the degree of personalization, and $w'_i$ is the adjusted tenant local model weight. This mechanism improves the adaptability of the model in diverse scenarios without affecting privacy and security and provides more accurate decision support for anomaly detection.

IV. EXPERIMENTAL RESULTS

A. Dataset

This study uses the Azure VM Telemetry Dataset as the primary data source. The dataset is collected from a large public cloud platform and is widely used in research on multi-tenant virtual machine (VM) monitoring and anomaly detection. It contains large-scale VM operation data over multiple periods. Metrics include CPU usage, memory consumption, disk I/O, network throughput, and service logs. The data has strong time-series characteristics and reflects the diverse behaviors of tenants. It provides a realistic view of dynamic resource usage in cloud environments.

Each entry in the dataset is associated with a timestamp and annotated with detailed metadata, including the specific resource type and instance identifier. This structure facilitates data partitioning and modeling at the tenant level. A subset of instances is further labeled with known system anomalies—such as resource overload, hardware failure, and malicious activity—providing a foundation for the development of supervised or semi-supervised anomaly detection models. In the context of federated learning, data subsets corresponding to distinct tenants can be treated as independent training clients. This configuration inherently aligns with the requirements of data distribution heterogeneity and privacy isolation in multi-tenant environments. The dataset is large in scale and broad in scope. It supports both global model training and personalized fine-tuning. This makes it suitable for evaluating the effectiveness of federated learning methods in distributed cloud computing settings. Its high-frequency sampling also helps capture short-term resource anomalies. This provides a solid foundation for building highly sensitive anomaly detection systems.

B. Experimental Results

Firstly, the results of the comparative experiment are given, as shown in Table 1.

Table 1. Comparative experimental results

| Method | Precision | Recall | F1-Score |
|---|---|---|---|
| Ours | 94.6 | 92.3 | 93.4 |
| DAGMM[22] | 87.1 | 89.5 | 88.3 |
| DeepLog[23] | 81.4 | 84.8 | 83.1 |
| RAD[24] | 90.3 | 87.0 | 88.6 |
| GDN[25] | 85.9 | 88.1 | 87.0 |

As illustrated in the comparison table, the proposed anomaly detection method based on federated learning exhibits superior performance across multiple evaluation metrics, including Precision, Recall, and F1-Score. Notably, it achieves an F1-Score of 93.4%, significantly outperforming all baseline methods. This result underscores the model's robust capability in accurately identifying anomalies within multi-tenant cloud environments, effectively mitigating both false positives and false negatives—factors critical to maintaining system stability and service quality. In contrast, conventional centralized approaches such as DAGMM and DeepLog demonstrate inferior performance, particularly in terms of Precision, likely due to their inadequate capacity to capture tenant-specific heterogeneity and the absence of inherent privacy-preserving mechanisms. More recent structure-aware models, including RAD and GDN, show improved adaptability to temporal and structural characteristics; however, their lack of federated coordination restricts their effectiveness in distributed scenarios, as evidenced by reduced recall. A detailed analysis of the impact of tenant participation on global performance is presented in Figure 2.

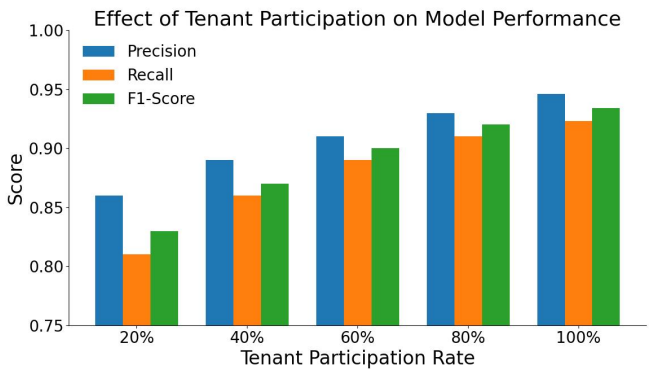

Figure 2. Analysis of the impact of multi-tenant participation ratio on global model performance

As shown in the figure, the model exhibits a consistent improvement in Precision, Recall, and F1-Score as the proportion of participating tenants increases, highlighting the positive impact of tenant involvement on global learning performance. Limited participation (e.g., 20%) constrains the model's ability to capture diverse anomaly patterns, whereas increased participation at 40–60% introduces valuable samples that enhance generalization, particularly improving Recall. At 80% and above, gains in Precision and F1-Score indicate improved discrimination between normal and abnormal behaviors, as federated aggregation more effectively mitigates data heterogeneity. Full participation (100%) yields the best overall performance, demonstrating the feasibility and effectiveness of the proposed federated learning approach in building a robust, privacy-preserving anomaly detection model for multi-tenant cloud environments. Additionally, the model's robustness under noise injection is evaluated, with results shown in Figure 3.

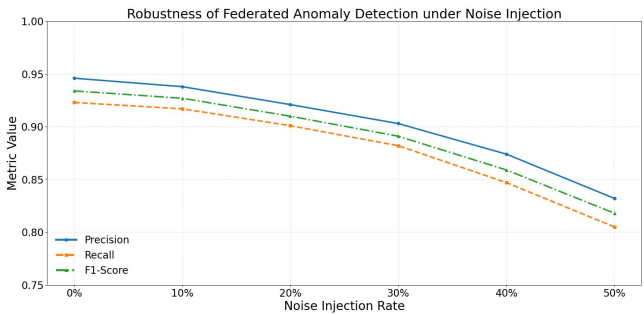

Figure 3. Impact of noise injection on the robustness of federated anomaly detection, measured across increasing noise injection rates.

As illustrated in the figure, the model exhibits a consistent decline in Precision, Recall, and F1-Score as the level of noise injection increases, indicating that noise adversely affects its ability to learn and detect anomalies. Despite this, performance remains relatively strong; for instance, with 50% noise, the F1-Score still reaches approximately 0.82, demonstrating the model's robustness. This resilience is largely attributed to the federated learning framework, which mitigates the impact of localized noisy data through collaborative aggregation. Precision shows only a modest decline, suggesting that the model continues to avoid excessive false positives, whereas Recall decreases more sharply, reflecting a reduced sensitivity to anomalies under high-noise conditions. These results underscore the importance of data quality in federated settings and suggest the value of integrating data reliability assessment and anomaly filtering into the learning process. Overall, the findings confirm the practical applicability and robustness of the proposed method, with the corresponding loss curve presented in Figure 4.

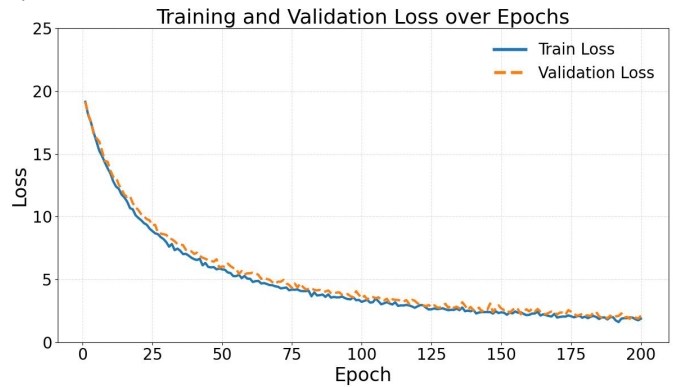

Figure 4. Training and validation loss over epochs, illustrating convergence and generalization performance.

As shown in the figure, both the training loss and validation loss exhibit a steady downward trend over 200 training rounds. This indicates that the model gradually learns the underlying patterns of tenant resource usage behavior during the training process. In the early stages, both curves drop rapidly, suggesting that the model quickly captures the dominant features of the anomaly detection task. This is a critical phase for building an effective detection model. It is worth noting that the validation loss curve shows slight local fluctuations, but the overall trend remains steadily decreasing. This suggests that the model is robust in handling heterogeneous tenant data and random disturbances. The stable downward trajectory reflects the strength of the federated aggregation mechanism in maintaining performance stability. In summary, the experiment not only shows the good convergence of the proposed model during training but also highlights its potential to build stable anomaly detection capabilities in dynamic multi-tenant resource scenarios. This lays a solid foundation for deploying a federated detection system in real-world cloud platforms.

## V. CONCLUSION

This study addresses the challenges of resource anomaly detection in multi-tenant cloud environments and proposes an intelligent detection method based on federated learning. The approach enables efficient and accurate anomaly identification while preserving data privacy. By introducing a federated learning framework, tenants can collaboratively train a global model without sharing raw data. This effectively tackles typical issues such as data heterogeneity, privacy isolation, and distributed computation. Experimental results show that the proposed method outperforms existing mainstream detection algorithms in several key metrics. It maintains high accuracy and strong robustness, offering solid support for building secure, reliable, and intelligent cloud resource management systems.

In the model design, data distribution differences in multi-tenant environments are fully considered. A local personalization mechanism is introduced on top of the global model to enhance adaptability to various tenant behavior patterns. To address common communication and performance bottlenecks in federated learning, hierarchical parameter aggregation, and personalized optimization strategies are applied. These strategies improve the stability and efficiency of the training process. The method shows strong practical applicability in complex scenarios involving dynamic resource usage and high-concurrency task scheduling. This demonstrates its feasibility for real-world deployment.

This research not only provides a theoretical foundation and practical approach for intelligent operations and system security in cloud computing but also promotes the application of federated learning in distributed system environments. Federated learning is increasingly needed in sensitive data domains such as finance, healthcare, and the Internet of Things. The proposed collaborative anomaly detection framework offers a transferable modeling paradigm for these fields. It is especially suited to complex tasks where data is decentralized, centralized training is not feasible, but high detection accuracy is required.

## VI. Future Work

Future work can explore asynchronous federated optimization, graph-based modeling mechanisms, and self-supervised learning strategies for anomaly detection. These directions may further improve the model's responsiveness and anomaly coverage. In addition, integrating edge computing can reduce detection latency, while incorporating multimodal monitoring data can enhance the model's semantic representation. Overall, the results of this study provide a strong technical foundation for building more intelligent, adaptive, and trustworthy cloud platforms. The work holds broad theoretical value and practical application potential.